 \documentclass[final,5p,times,twocolumn]{elsarticle}

\usepackage{amssymb}

\usepackage{amsmath,amsfonts}
\usepackage{algorithmic}
\usepackage{graphicx}
\usepackage{textcomp}
\usepackage{xcolor}

\usepackage{algorithm}
\usepackage{algorithmic}
\usepackage{subfigure}
\usepackage{epstopdf}
\usepackage{enumerate}
\usepackage{tabularx}
\usepackage{amsthm}
\usepackage{booktabs}
\usepackage{balance}
\usepackage{xspace}
\usepackage{multirow}

\newcommand{\ourmethod}{\texttt{GST-Pro}\xspace}
\newcommand{\mts}{multivariate time series\xspace}
\newcommand{\mtsad}{multivariate time series anomaly detection\xspace}

\usepackage{soul}
\usepackage{xcolor}

\newcommand{\revision}[1]{{\textcolor{black}{#1}}}
\usepackage{url}

\journal{Information Fusion}

\begin{document}

\begin{frontmatter}



\title{Graph Spatiotemporal Process for Multivariate Time Series Anomaly Detection with Missing Values}

\author[label1]{Yu Zheng\fnref{eq}}\ead{yu.zheng@latrobe.edu.au}

\author[label2]{Huan Yee Koh\fnref{eq}}\ead{huan.koh@monash.edu}

\author[label2]{Ming Jin}\ead{ming.jin@monash.edu}

\author[label1]{Lianhua Chi\corref{cor}}\ead{l.chi@latrobe.edu.au}

\author[label4]{Haishuai Wang}\ead{haishuai.wang@zju.edu.cn}

\author[label1]{Khoa T. Phan}\ead{k.phan@latrobe.edu.au}

\author[label1]{Yi-Ping Phoebe Chen}\ead{phoebe.chen@latrobe.edu.au}

\author[label3]{Shirui Pan}\ead{s.pan@griffith.edu.au}

\author[label1]{Wei Xiang}\ead{W.Xiang@latrobe.edu.au}

\affiliation[label1]{organization= {Department of Computer Science and Information Technology, La Trobe University},
            country={Australia}}
\affiliation[label2]{organization={Department of Data Science and AI, Faculty of IT, Monash University},
            country={Australia}}
\affiliation[label3]{organization= {School of Information and Communication Technology, Griffith University},
            country={Australia}}
\affiliation[label4]{organization= {Zhejiang University},
            country={China}}
\fntext[eq]{Y. Zheng and H. Y. Koh contributed equally to this work.}
\cortext[cor]{L. Chi is the corresponding author.}


\begin{abstract}
The detection of anomalies in multivariate time series data is crucial for various practical applications, including smart power grids, traffic flow forecasting, and industrial process control. However, real-world time series data is usually not well-structured, posting significant challenges to existing approaches: (1) The existence of missing values in multivariate time series data along variable and time dimensions hinders the effective modeling of interwoven spatial and temporal dependencies, resulting in important patterns being overlooked during model training; (2) Anomaly scoring with irregularly-sampled observations is less explored, making it difficult to use existing detectors for multivariate series without fully-observed values. In this work, we introduce a novel framework called \ourmethod, which utilizes a \underline{g}raph \underline{s}patio\underline{t}emporal \underline{pro}cess and anomaly scorer to tackle the aforementioned challenges in detecting anomalies on irregularly-sampled multivariate time series. Our approach comprises two main components. First, we propose a graph spatiotemporal process based on neural controlled differential equations. This process enables effective modeling of multivariate time series from both spatial and temporal perspectives, even when the data contains missing values. Second, we present a novel distribution-based anomaly scoring mechanism that alleviates the reliance on complete uniform observations. By analyzing the predictions of the graph spatiotemporal process, our approach allows anomalies to be easily detected. Our experimental results show that the \ourmethod method can effectively detect anomalies in time series data and outperforms state-of-the-art methods, regardless of whether there are missing values present in the data. Our code is available: \url{https://github.com/huankoh/GST-Pro}
\end{abstract}




\end{frontmatter}


\section{Introduction}\label{sec:intro}


Swift technological advancement has brought an explosive surge in the pervasiveness and volume of time series data. Ranging from health care \cite{lee2017big} and critical infrastructures \cite{su2019robust,li2021multivariate,mathur2016swat} to spacecrafts \cite{hundman2018detecting}, various industries now generate data from numerous devices or sensors across time, forming a complex multivariate time series with hundreds to thousands of variables. This surge in multivariate time series data has inevitably led us to place significant reliance on the automatic detection of anomalous events through multivariate time series data to identify, avert, and respond to catastrophic events before and as they occur. Ideally, the detection can be done through algorithms that can be implemented at scale, and are robust to the noise due to imperfections of practical real-world systems. Consequently, there has been a strong demand for the development of robust \mts anomaly detection models \cite{garg2021evaluation,schmidl2022anomaly,darban2022deep}.  

Although there is an abundance of multivariate time series data that exhibit normal patterns, anomaly events are typically associated with rare events, so collecting and labeling anomalies are often a daunting task. As a result, unsupervised anomaly detection techniques have been widely explored as a practical solution to the challenging anomaly detection problem. Amongst the proposed techniques, the classical methods include statistical unsupervised models such as ARIMA/VAR \cite{yu2016improved}, 
 distance-based \cite{keogh2005hot} or distributional \cite{ting2021isolation} approaches. However, these methods may have limitations in capturing the non-linear spatial and temporal relationships present in multivariate time series data \cite{garg2021evaluation}.
 
More recently, with the flourishing of deep learning (DL), significant advances have been made. Early work from Hundman et al. \cite{hundman2018detecting} proposed a Long Short-Term Memory (LSTM) network to detect anomalies based on the forecasting errors, and Park et al. \cite{park2018multimodal} proposed a reconstruction-based LSTM framework based on the reconstruction errors. Nonetheless, LSTM frameworks \cite{su2019robust} lack explicit modeling of pairwise inter-dependence among variable pairs, which limits their ability to detect complex anomaly events in high-dimensional multivariate time series data\cite{zhao2020multivariate}. In response, MTAD-GAT \cite{zhao2020multivariate} and GDN \cite{deng2021graph} use spatiotemporal graph neural networks (STGNNs) to model spatial and temporal data correlations. GDN, in particular, uses a graph learning layer to learn pairwise correlations, negating the need for a predefined graph, often unavailable in multivariate time series datasets. To date, STGNNs remain as the state-of-the-art anomaly detection models for multivariate time series \cite{han2022learning}.

Despite significant advancements in multivariate time series anomaly detection, existing deep anomaly detection techniques rely on well-structured, \textit{regular} time series data that is \textit{sampled at a uniform frequency}. However, real-world multivariate time series data often has random missing values \cite{little2019statistical} or non-uniform observations due to \textit{irregular} sampling frequencies. 
Random missing values in multivariate time series data are usually due to sensor limitations or transmission interruption, while non-random missing values can be caused by multi-modal sources and process heterogeneity \cite{mitra2023learning}. Even for small-scale systems, random missing values are almost unavoidable. Hence, at the very least,  it is critical to develop robust techniques that can accurately detect anomalous events despite such imperfections. Nonetheless, the irregular multivariate time series anomaly detection problem has not been a well-investigated setting thus far. 

To address missing data points, straightforward solutions such as zero padding, interpolation and imputation algorithms \cite{beretta2016nearest}, and linear predictors \cite{durbin2012time} can be utilized. This way, the missing value problem in the multivariate time series can be simply resolved as a pre-processing step. However, as will be demonstrated in our experimental results, using a modular pipeline of impute-then-detect approach can lead to significantly weakened anomaly detection performances. This necessitates an alternative approach to detect anomalous events in irregular multivariate time series.

\begin{figure*}[t]
    \centering
       \includegraphics[width=.8\textwidth]{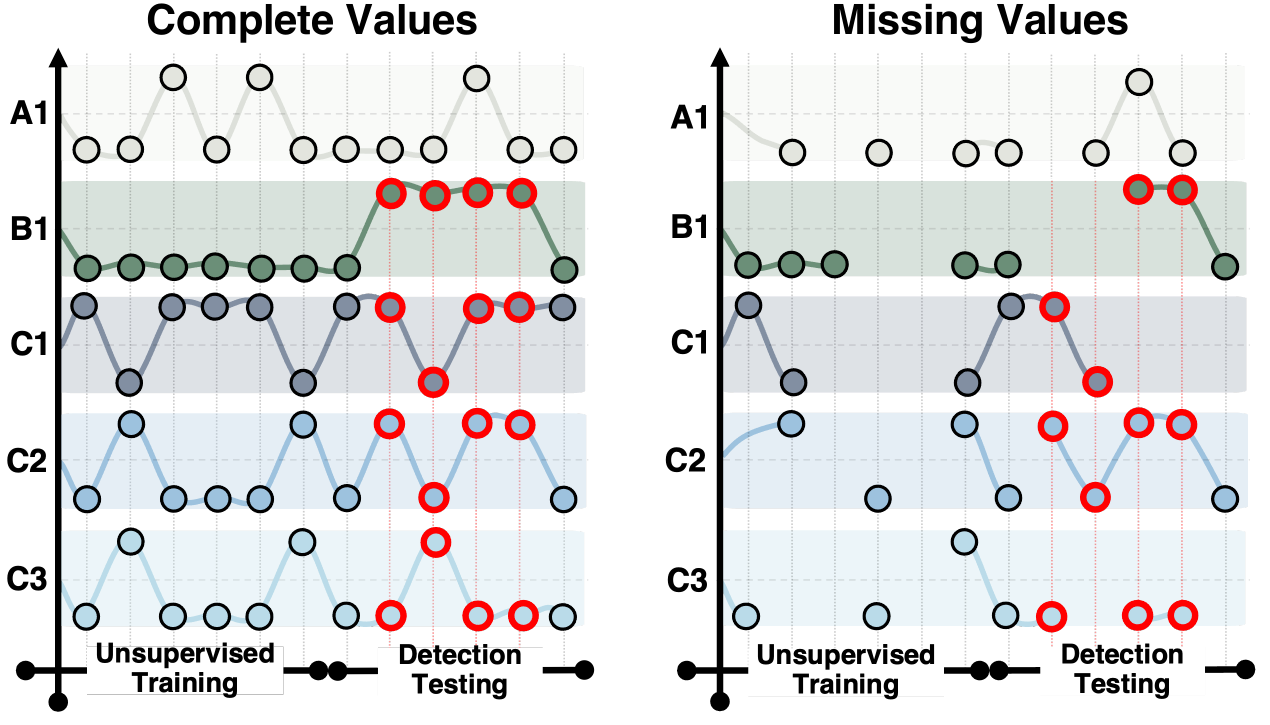}\\
       {\caption{A simplified comparison of multivariate time series data with complete values (left) and missing values (right). First 6 observations are the normal data used for unsupervised training and the last 6 are to test a model in detecting anomaly events. Red circles indicate that it is an anomalous observation as manifested by four middle observations in the complete test data.}\label{fig. intro}}

\end{figure*}

The difficulties of detecting anomaly events in multivariate time series with missing values can be illustrated in Figure \ref{fig. intro}. Following standard unsupervised anomaly detection setting, the first 6 observations of normal data is used for training models, while the testing data contains normal (first and last) and anomalous (four middle) timestamps that are used to evaluate models in detecting the anomalies. 

Firstly, A1 series has no anomalous event, but will records a short burst of values intermittently. While this can be easily learned in a regular series setting, irregular multivariate time series may not record the short burst during normal training periods and models may think these short burst events are anomalous during detection testing. In short, under missing values setting,  \textit{high quality training data for unsupervised learning is sparse}. 

Secondly, as shown in right plot B1, the real observed values may be inaccessible during anomalous period, making it hard to detect anomalies in real-time streaming data. Hence, \textit{real observed values cannot be safely relied upon to detect anomaly events in real time}. 

Both A1 and B1 show a time series that record values that move independently. However, in multivariate time series, the variables are intricately related. For example, C1, C2, and C3 are interrelated variables where C1 has a strong negative correlation with C2, while C3 have a strong positive correlation with C2. A deviation from these relationships is thus an anomaly event. In right plot of Figure \ref{fig. intro}, we see C1 recorded conflicting values with C2 only for first two early timepoints, but not the subsequent anomalous values. In contrast, C3 records the anomalous values in the later periods rather than the early periods. As it is very unlikely that all channels have missing values concurrently under a random missing scenario, we conjecture that if a model fully captures the inter-dependence between the variables, it can detect anomaly events well in irregular \mts even if the data suffers from high missing rate. Hence, it is crucial for an anomaly detection model to \textit{explicitly capture the complex pairwise associated relationship (i.e., degree of spatial dependence) between the variables of a multivariate time series}. 

Based on the above observations, we summarize the challenges for irregular multivariate time series anomaly detection:
\begin{itemize}
\item \textbf{Sparsity in high quality training data (Challenge 1)} The presence of irregularity in multivariate time series can lead to important spatio-temporal patterns being omitted during the training phase of model development.
\item \textbf{Anomaly scoring with irregular observations (Challenge 2)} A model should enable real-time detection of anomaly events despite the inability to ensure complete access to observed values in a \mts. 
\item \textbf{Spatial-temporal dependency modeling (Challenge 3)} Multivariate time series analysis requires a deep understanding of spatial-temporal dependency; how to simultaneously capture spatial and temporal dependency given the missing values problem is the ultimate challenge for \mtsad. 
\end{itemize}

To address these challenges, we propose a novel prediction-based anomaly scorer that leverages our graph spatiotemporal processes to model multivariate time series, whether they contain missing values or not. Specifically, our approach involves imputing the missing values in each variable of the input multivariate time series to generate a set of continuous paths. We then design two neural controlled differential equation (NCDE) processes to model the input data from both spatial and temporal perspectives, addressing the first and third challenges mentioned above at once. By incorporating these processes, we are able to model any multivariate time series, regardless of whether they contain missing values. To address the second challenge, we propose a novel distribution-based anomaly scorer that is built on top of our time series model, providing two significant advantages: (1) It is solely based on model predictions and does not require comparisons with ground truths, avoiding issues arising from missing values when calculating real-time anomaly scores; (2) It is based solely on prediction statistics and does not contain trainable parameters, making it a plug-and-play module that can be even integrated with other time series models beyond our method. By combining the forecaster and anomaly scorer discussed above, our proposed \ourmethod method, as shown in Figure \ref{fig. framework}, can effectively detect anomalies in arbitrary real-world multivariate time series data in an online and unsupervised manner.
The main contributions of this paper are as follows:
\begin{itemize}
\item We propose dynamic graph neural differential equations (DG-NCDEs) to model multivariate time series, particularly those with missing values.
\item We propose a parameter-free anomaly detector for multivariate time series data, built on top of our forecasting model, that can detect anomalies in an online and unsupervised manner.
\item We conduct extensive experiments comparing \ourmethod with state-of-the-art baselines under various settings, demonstrating the superiority of our method.
\end{itemize}

The rest of the paper is structured as follows. Section~\ref{sec:rw} reviews the related work. Section~\ref{sec:definition} gives the definition of the task. Section~\ref{sec:model} presents the proposed \ourmethod. Section~\ref{sec:experiments} illustrates our experiments and followed by conclusion in Section~\ref{sec:conclusion}.

\section{Related Work}\label{sec:rw}
In this section, we introduce the related works on unsupervised time series anomaly detection and spatiotemporal graph neural networks. 

\subsection{Unsupervised Time Series Anomaly Detection}\label{subsec:rw_mts_ad}
Prior literature on unsupervised time series anomaly detection can be broadly categorized as forecasting-based \cite{yu2016improved}, reconstruction-based \cite{zhao2020multivariate}, distance-based \cite{keogh2005hot} and distribution-based \cite{ahmad2017unsupervised} methods. 

Forecasting-based methods fundamentally rely on forecasting errors to detect anomalies. Particularly, after optimizing a model on normal training data, the model will typically predict the one-step ahead forecast for current timestamp. The forecast values are then compared to the observed values at current timestamp to determine how anomalous the current timestamp is. As such, many classical forecasting models such as ARIMA/VAR \cite{yu2016improved} can be adapted for this purpose \cite{schmidl2022anomaly}. Early DL works introduced recurrent neural networks (RNN)\cite{hundman2018detecting} and, lately, Transformers\cite{song2018attend,chen2021learning} to scale through high dimensional data and model complex non-linear patterns\cite{garg2021evaluation}. 

Reconstruction-based methods detect anomalies based on reconstruction errors. Conceptually, this involves learning the representation of normal training series and outputs a lossy reconstruction of the input. As the learned representation is optimized for normal data, a high reconstruction error is likely during anomalous periods. Classical reconstructions methods include PCA\cite{garg2021evaluation} and AutoEncoder \cite{zhang2019deep}. Similarly, numerous deep reconstruction models have also been proposed to model the complex spatiotemporal dependencies of multivariate time series, including AutoEncoder (AE) \cite{zhang2019deep}, Variational AutoEncoder (VAE) \cite{park2018multimodal, su2019robust,li2021multivariate}, Normalizing flows \cite{dai2022graphaugmented}. 

Distance-based methods utilize specialized metrics to compare points or subsequences of a multivariate time series with each other, including kNN and local outlier factor \cite{breunig2000lof}. On the other hand, distribution approaches detect anomalies by judging the likelihood of an observation after fitting a distribution model to windowed points or subsequences of the time series \cite{ting2022new}. With a few exceptions of distance-based approaches \cite{shen2020timeseries,shin2020itad}, DL architectures are mostly characterized by the reconstruction or forecasting approaches as detailed in the most recent survey \cite{darban2022deep}. 

\subsection{Spatial-Temporal Graph Neural Networks}
GNNs \cite{zhang2022trustworthy} have recently become de facto models to analyse graph data such as social and academic network~\cite{zhang2023demystifying,zhang2023interaction}, drug discovery \cite{koh2023psichic,nguyen2023gpcr,zheng2023large} and natural language~\cite{koh2022empirical,koh2022far}. To handle dynamic data, GNNs have been exploited under the umbrella of \textit{spatial-temporal graph neural networks} (STGNNs) \cite{seo2018structured,jin2023survey}. As STGNNs can explicitly model fine-grained spatial associations between the multivariate series channels, they are particularly well suited for scenarios where the underlying graph structure remains fixed but the node features dynamically update over time \cite{wu2020connecting,zheng2023correlation}. Early work from \cite{seo2018structured} proposed a recurrent-based STGNN for forecasting multivariate time series by capturing the temporal dependency using RNN and spatial relation using GNN. Since then the STGNNs field has flourished with architectures proposed to tackle various time series applications including forecasting \cite{jin2022multivariate} and classification \cite{duan2022multivariate}. For multivariate time series anomaly detection, Deng and Hooi \cite{deng2021graph} recently proposed GDN that leverages STGNN to make a one-step-ahead forecast and detects anomalies by computing the normalized forecasting errors. MTAD-GAT \cite{li2021multivariate}, utilizes a joint forecasting and reconstruction-based STGNN to detect anomalies. FuSAGNet extends on MTAD-GAT with a sparsity-constrained joint optimization of STGNN \cite{han2022learning}.

In spite of the significant advancements, current deep detection techniques require real observations to be completely available at every timestamp, which is not possible under irregular time series settings. In our work, we leverage STGNN with DG-NCDE spatial and temporal modules to handle irregular time series, and propose a simple distributional approach on original forecasts of \ourmethod rather than forecasting errors to achieve robust anomaly detection in real-time. 

\section{Problem Formulation}\label{sec:definition}
A multivariate time series with successive, equal-spaced observations can be defined as $\mathbf{X} =\{\mathbf{x}^\textit{(1)}, \mathbf{x}^\textit{(2)}, \cdots, \mathbf{x}^\textit{(T)}\}$ where $T$ represents the length of the series at current timestamp and $\mathbf{x}^{(t)} \in \mathbb{R}^{N}$ is composed of $N$ number of univariate channels $\{x^{(t)}_{1},x^{(t)}_{2}, \cdots,x^{(t)}_{N}\}$. In our work, we focus on real-time, multi-modal sensing data that consists of $N$ sensor nodes over $T$ timestamps. Hence, each univariate channel is also referred to as ``sensor" or ``node" interchangeably. 

For a \textit{regular} unsupervised \mts anomaly detection, we are required to learn an anomaly classifier or scorer, $A(\cdot)$, that outputs an anomaly score to each timestamp that clearly differentiates anomalous observations and non-anomalous observations. Particularly, the outputs can be conceptualized as an indicator that informs a system operator whether the timestamp is anomalous or not, $A(\mathbf{x}^{(a)}) > A(\mathbf{x}^{(n)})$, where $\mathbf{x}^{(a)}$ is anomalous observation and $\mathbf{x}^{(n)}$ is not. The ground truth label that indicates whether a timestamp is anomalous or not is represented as $g^{(t)} \in \{0,1\}$ and $g^{(t)} = 1$ if the observation $\mathbf{x}^{(t)}$ is anomalous. As the detection of anomalies takes place when the data is streamed in real-time, models can only rely on past observations to make a decision at every timestamp and cannot reverse their previous decisions.

For an \textit{irregular} unsupervised \mts anomaly detection, we aim to achieve the same goal but under the presence of missing values in the data. In this study, we consider the missing-at-random scenario \cite{rubin1976inference}. The missing values issue is present in the training data used for unsupervised training and testing data used for evaluating model detection of anomaly events. Such irregularities are common for practical real-world \mts, and thus is the focus of this work.  

\section{Methodology}\label{sec:model}
In this section, we explain the architecture of our proposed method, \ourmethod, including two key components: the DG-NCDE-based forecasting head (Section \ref{sec:forecasting module}), and the Gaussian scoring-based anomaly detector (Section \ref{sec:ano_scorer}). We begin with the overall architecture design in Section \ref{sec:overall architecture}.
\begin{figure*}[!hbt]
    \centering
       \includegraphics[width=1\textwidth]{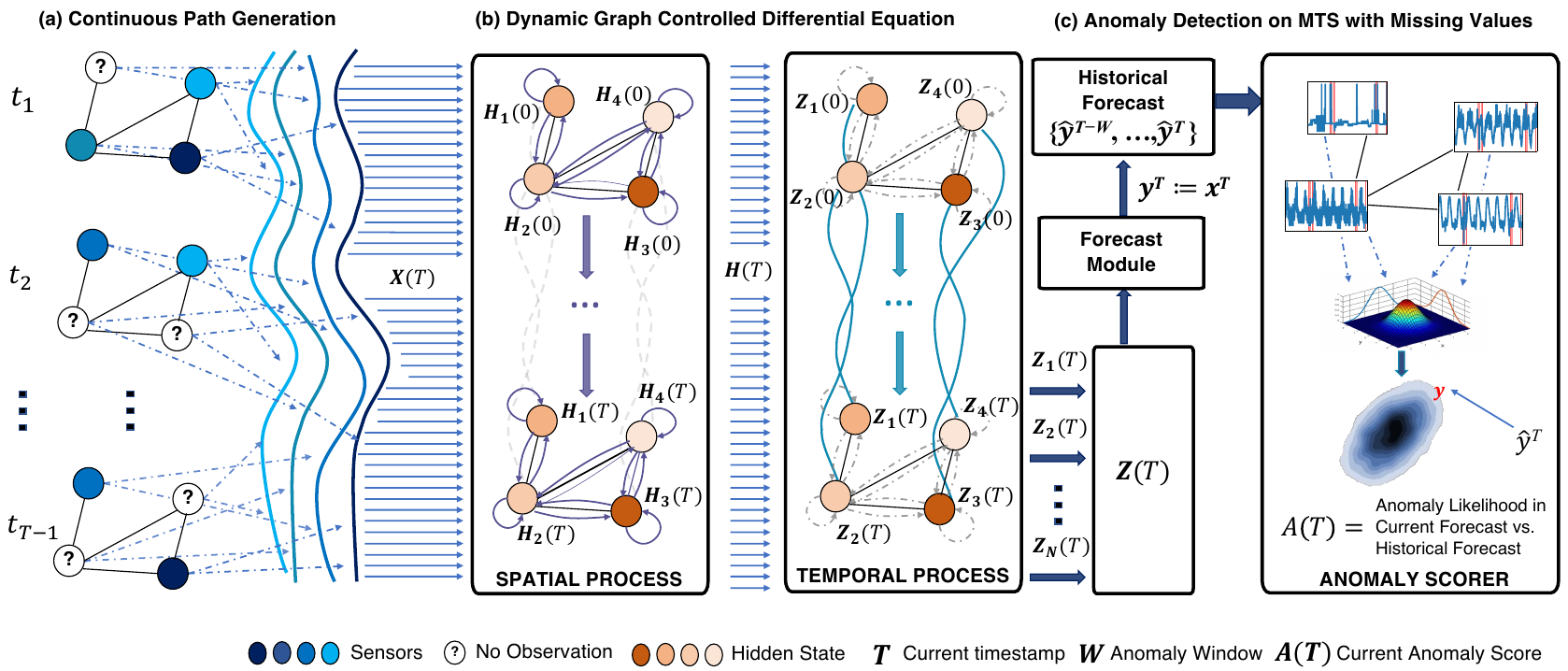}\\
       {
       \caption{
       Overall Framework of \ourmethod. \textbf{(a)}: Imputing discrete observations with missing values to generate continuous paths. \textbf{(b)}: Utilizing two neural controlled differential equations to process input signals from both spatial and temporal perspectives. \textbf{(c)}: Conducting real-time anomaly detection on arbitrary multivariate time series \revision{using a distribution-based scorer that assesses anomalies based on forecast outputs, without ground-truth observations. Anomaly scores are calculated by evaluating the likelihood of deviations in current forecasts from historical forecasts using a multivariate normal distribution.}
       }\label{fig. framework}}
\end{figure*}

\subsection{Overall Architecture}\label{sec:overall architecture} 
Our approach, \ourmethod, as illustrated in Fig. \ref{fig. framework}, detects anomalies in \mts with missing values. Initially, \ourmethod imputes missing values to form continuous paths. Then, taking the imputed set of continuous paths, \ourmethod employs spatial and temporal neural controlled differential equation processes to optimize the forecasting module in making a one-step-ahead forecast. We posit that by optimizing these processes for forecasting, the model can effectively discern non-anomalous spatiotemporal dependencies from normal training data. 

When there are no anomalies, the module is expected to produce normal forecasts that resembles the values in the non-anomalous training data. Conversely, during periods of anomalies, the module will generate outputs that deviate significantly from the normal forecasts. 

Building on this concept,  we propose an anomaly scorer that evaluates the abnormality of the forecast values by estimating the probability of anomalies at each timestamp. Hence, the irregularly observed values are used exclusively for one-step-ahead forecasting, while the anomaly score is computed by the scorer solely from historical and current forecasts.

\subsection{Dynamic Graph Neural Controlled Differential Equations}\label{sec:forecasting module}
A multivariate time series is typically conceptualized as a discrete-time dynamic graph composed of regularly-sampled graph snapshots, denoted as ${\mathcal{G}_t := (\mathcal{V}, \mathcal{E}, \mathbf{x}^{(t)})}$. Here, $\mathcal{V}$ and $\mathcal{E}$ define the predetermined graph structure for a sequence of observations, characterizing the underlying connectivity between variables (sensors), while $\mathbf{x}^{(t)}$ refers to the node features of the snapshot $\mathcal{G}_t$ at time $t$. However, in practice, missing values can be present in both the variable and time dimensions due to data sampling or sensor failures, posing a significant challenge for utilizing off-the-shelf dynamic graph neural networks, such as MTGODE \cite{jin2022multivariate}, as the model for embedding multivariate time series. In this work, we follow \cite{choi2022graph} and address this challenge by proposing the Dynamic Graph Neural Controlled Differential Equations (DG-NCDE) to model multivariate time series, regardless of the presence of missing values in the data. The formulation of Neural Controlled Differential Equations (NCDEs) \cite{kidger2020neural} is shown in Eq. \ref{eq: NCDE}, which is built on the basis of Neural Ordinary Differential Equations (NODEs) \cite{chen2018neural}, and the end goal is to learn a CDE function $f(\cdot; \mathbf{\Theta})$, parameterized by $\mathbf{\Theta}$, from the data.
\begin{align}
    \mathbf{Z}(\tau) &= \mathbf{Z}(t_0) + \int_{t_0}^{\tau} f(\mathbf{Z}(t); \mathbf{\Theta}) d\Tilde{\mathbf{X}}(t), \nonumber \\
    &= \mathbf{Z}(t_0) + \int_{t_0}^{\tau} f(\mathbf{Z}(t); \mathbf{\Theta}) \frac{d\Tilde{\mathbf{X}}(t)}{dt} dt.
    \label{eq: NCDE}
\end{align}
In the above equation, the evolution of the hidden state $\mathbf{Z}(t)$ is controlled over time based on $\Tilde{\mathbf{X}}(t)$, which denotes a continuous path derived from the discrete observations $\mathbf{X}$. In this regard, NCDEs can be viewed as a continuous version of Recurrent Neural Networks (RNNs) \cite{kidger2020neural} and demonstrates superior performance in many time series benchmarks \cite{choi2022graph}. In addition to the effectiveness of NCDEs, a notable advantage of Eq. \ref{eq: NCDE} is its flexibility in handling input data, as it does not impose strong constraints such as no missing values. Though NCDEs shed light on modeling real-world univariate time series, it remains unclear how to model (discrete-time) dynamic graphs with NCDEs. To address this gap and inspired by \cite{sankar2020dysat} and \cite{choi2022graph}, we propose two different processes to model the entangled spatial and temporal dependencies in the input data (detailed in subsections \ref{sec:spatial process} and \ref{sec:temporal process}). In a nutshell, we model a multivariate time series by solving the following equation that combines spatial and temporal processes together.
\begin{equation}
    \mathbf{Z}(\tau) = \mathbf{Z}(t_0) + \int_{t_0}^{\tau} f(\mathbf{Z}(t); \mathbf{\Theta}) g(\mathbf{H}(t); \mathbf{\Phi}) \frac{d\Tilde{\mathbf{X}}(t)}{dt} dt.
    \label{eq:DG-NCDE}
\end{equation}
Here, $g(\cdot; \mathbf{\Phi})$ and $f(\cdot; \mathbf{\Theta})$ denote spatial and temporal NCDE functions, each parameterized by distinct parameter sets, designed to model the inherent spatial and temporal dynamics of the input data, respectively. Our formulation presented in Eq. \ref{eq:DG-NCDE} can be conceptualized as a continuous-time approach to modeling a discrete-time dynamic graph, similar to the method proposed in \cite{jin2022multivariate}.  However, the distinguishing feature of DG-NCDE is its capability to model multivariate time series data, whether or not missing values are present. To evaluate DG-NCDE and calculate the gradients of our method, $\Tilde{\mathbf{X}}(t)$ has to be twice continuously differentiable as same as in NCDEs \cite{kidger2020neural}. In practice, we first process the original series into sliding window inputs with length $w_s$, and use natural cubic spline interpolation \cite{mckinley1998cubic} to generate the continuous path $\Tilde{\mathbf{X}}(t) \in \mathbb{R}^{N \times w_s}$ for each window as a pre-processing stage based on the available discrete observations. The sliding window input includes observations $\mathbf{x}^{(t)}$ where $t \in \{T-w_s, \cdots, T-1\}$ for the forecasting module to make a one-step-ahead forecast. In the subsequent two subsections, we elucidate Eq. \ref{eq:DG-NCDE} through the lenses of spatial and temporal processes, respectively.

\subsubsection{Spatial Process}\label{sec:spatial process}
We first model the hidden state trajectory of each variable between graph snapshots from the perspective of message passing controlled by the continuous path $\Tilde{\mathbf{X}}(t)$. 

Formally, we define the spatial NCDE as follows with the bounds $t_0 := T-w_s$ and $\tau := T-1$.
\begin{equation}
    \mathbf{H}(\tau) = \mathbf{H}(t_0) + \int_{t_0}^{\tau} g(\mathbf{H}(t); \mathbf{\Phi}) \frac{d\Tilde{\mathbf{X}}(t)}{dt} dt.
    \label{eq:spatial NCDE}
\end{equation}
In terms of the formulation of spatial NCDE function $g(\mathbf{H}(t); \mathbf{\Phi})$, we approximate the graph convolution with the first-order Chebyshev polynomial as demonstrated in \cite{kipf2016semi}. 
\begin{equation}
    g(\mathbf{H}(t); \mathbf{\Phi}) = \text{FC}^{(1)}\big(\Hat{\mathbf{D}}^{-\frac{1}{2}}\Hat{\textbf{A}}\Hat{\mathbf{D}}^{-\frac{1}{2}} \Tilde{\textbf{H}}(t) \mathbf{W}_s\big),
    \label{eq:spatial NCDE function}
\end{equation}
where $\Tilde{\textbf{H}}(t) = \text{ReLU}\big(\text{FC}^{(0)}\big(\textbf{H}(t)\big)\big)$, $\Hat{\textbf{A}} = \mathbf{A} + \mathbf{I}$, and $\Hat{\mathbf{D}}$ denotes the diagonal degree matrix of $\Hat{\textbf{A}}$. For simplicity, we let $\mathbf{\Phi}$ denote the set of trainable parameters, e.g., $\mathbf{W}_s$ and the parameters in fully-connection layers $\text{FC}^{(0)}(\cdot)$ and $\text{FC}^{(1)}(\cdot)$, in the above spatial NCDE function. In the case of a predefined graph adjacency matrix $\mathbf{A}$ is unavailable, we learn it end-to-end with the entire model \cite{choi2022graph}, i.e., $\textbf{A} := \text{ReLU}(\mathbf{E} \cdot \mathbf{E}^\top)$, where $\mathbf{E}$ denotes a trainable node embedding matrix.

\subsubsection{Temporal Process}\label{sec:temporal process}
To learn from temporal dependencies, we introduce another process, known as temporal NCDE, that explicitly models temporal patterns. Formally, we define this process as follows with the same bounds as in Eq. \ref{eq:spatial NCDE}.
\begin{equation}
    \mathbf{Z}(\tau) = \mathbf{Z}(t_0) + \int_{t_0}^{\tau} f(\mathbf{Z}(t); \mathbf{\Theta}) \frac{d\mathbf{H}(t)}{dt} dt.
    \label{eq:temporal NCDE}
\end{equation}
In the above formulation, \revision{$f(\mathbf{Z}(t); \mathbf{\Theta})$ is the temporal NCDE function, where the evolution of hidden trajectories $\mathbf{Z}(t)$ across time is controlled by the continuous path $\mathbf{H}(t)$ that generated by the spatial process.} There exists a wide range of implementations of $f(\mathbf{Z}(t); \mathbf{\Theta})$. For this study, we have chosen the method presented in \cite{choi2022graph}, \revision{which involves modeling each trajectory, i.e., each column of $\mathbf{Z}(t)$, with individual fully-connected layers.}
\begin{equation}
    f(\mathbf{Z}(t); \mathbf{\Theta}) = \phi(\text{FC}^{(0)}(\mathbf{Z}_0(t)))\ ||\  \cdots\ ||\ \phi(\text{FC}^{(N-1)}(\mathbf{Z}_{N-1}(t))),
    \label{eq:temporal NCDE function}
\end{equation}
where $\phi(\cdot)$ and $||$ denote ReLU activation and the operation of concatenation, respectively. Similar in Eq. \ref{eq:spatial NCDE} and for simplicity, we use $\mathbf{\Theta}$ to represent the set of trainable parameters in our temporal NCDE function.
\revision{Intriguingly, with the configuration illustrated in Eq. \ref{eq:temporal NCDE function}, we have Eq. \ref{eq:temporal NCDE} transformed into a continuous RNN, which effectively captures the inherent temporal dependencies.}

Our formulation of DG-NCDE, as defined in Eq. \ref{eq:DG-NCDE}, combines the spatial and temporal processes to model a given multivariate time series. This framework allows for the forecasting of future data points by passing the learned time series representations through an additional fully-connected layer, acting as the downstream forecaster. With forecast output for each timestamp denoted as $\Hat{\mathbf{y}}^{(t)}$, we provide a detailed explanation of our model training process in Section \ref{sec:model training}.

\subsection{Anomaly Scoring}\label{sec:ano_scorer}
To design a robust anomaly scorer that handles time series with missing values, \ourmethod assesses the abnormality at each timestamp without the need of accessing real observed values at the current timestamp. Specifically, at current timestamp, $T$, the only required inputs for the anomaly scorer of \ourmethod are the current and historical forecasts, $ \{\Hat{\mathbf{y}}^{(1)}, \cdots, \Hat{\mathbf{y}}^{(T)}\}$ from forecasting module. 

For each channel, we first compute its \textit{anomaly likelihood}, $\alpha^{(t_i)}$, of the current forecast by computing negative log-likelihood of the one-step-ahead forecast value after fitting a rolling Gaussian distribution on past and current forecast values:
\begin{equation}     \alpha^{(t)}_i(\hat{y}^{(t)}_i|\hat{\mu}^{(t)}_i,\hat{\sigma}^{(t)}_i) = \ln \hat{\sigma}^{(t)}_i + \frac{1}{2}\ln(2\pi) + \frac{1}{2}\left( \frac{\hat{y}^{(t)}_i - \hat{\mu}^{(t)}_i}{\hat{\sigma}^{(t)}_i}\right)^2      
\label{eq: ano_likelihood}
\end{equation}
where $\hat{\mu}^t_i$ and $\hat{\sigma}^t_i$ represent the mean and standard deviation parameter of our rolling Gaussian distribution:
\begin{equation}
    \hat{\mu}^{(t)}_i = \frac{1}{W} \sum_{j=0}^{W-1} \hat{y}_i^{(t-j)} \;\text{;} \;(\hat{\sigma}^{(t)}_i)^2 = \frac{1}{W} \sum_{j=0}^{W-1} \left(\hat{y}_i^{(t-j)} -\hat{\mu}^{(t)}_i \right)^2
\label{eq: gaus_para}
\end{equation}
with $W$ being the anomaly window size. Following \cite{garg2021evaluation}, we prepend the last $W - 1$ values from the training data to calculate the Gaussian parameters, $\hat{\mu}^{(t)}_i$ and $\hat{\sigma}^{(t)}_i$, for $ t < W$.

Intuitively, this gives us the anomaly likelihood at the current timestamp by determining how anomalous the current timestamp forecast is compared to the historical forecast of \ourmethod. Hence, the points that are at the tail of the distribution would then have high anomaly scores. To compute the final anomaly score, $A^{(t)}$, for each timestamp, we linearly aggregate the anomaly likelihood for each channel:
\begin{equation}
    A^{(t)} = \sum^{M}_{i=1} \alpha^{(t)}_i
\label{eq: ano_score}
\end{equation} 

Anomaly scorer of \ourmethod is similar to the Gaussian scorers in \cite{ahmad2017unsupervised,garg2021evaluation}, but differs in that we only take the forecast outputs rather than the reconstruction or forecast error. In fact, state-of-the-art approaches \cite{su2019robust,deng2021graph,garg2021evaluation} primarily rely on observed values to compute anomaly score in the form of reconstruction or forecasting error. While forecasting errors allow simple detection of anomaly events based on the deviation between predicted and real values, the real observations cannot be relied upon safely under an irregular time series scenario as they are constantly missing and inaccessible. 

In contrast, our approach relies on the assumption that \ourmethod can predict values that closely resemble forecast values during the non-anomalous period, but generate outputs that degenerate and deviate from normal forecasting outputs during anomalous periods. In other words, as long as there is spatial and/or temporal abnormality in the signals of the input sliding window, \ourmethod should generate forecasts that are also anomalous. Conversely, it should generate forecast values that are similar to forecast outputs made for non-anomalous training data if there are no anomalous signals in the input. 

This approach frees us from making any assumptions, which are required to accurately impute the missing values. Rather, we place importance on the forecasting module of \ourmethod to learn the normality of spatial and temporary dependency from training data. To be demonstrated in the next section, we argue that this assumption is not only valid but desirable as we see a minor performance drop under even high missing rate scenarios.  

\subsection{Model Training}\label{sec:model training}
While our anomaly detector does not contain trainable parameters, unsupervised training is required for the proposed DG-NCDE forecasting module. To achieve this, we follow the methodology outlined in \cite{choi2022graph} by constructing the augmented ODE below instead of individually implementing Eq. \ref{eq:spatial NCDE} and \ref{eq:temporal NCDE}:

\begin{equation}
\frac{d}{dt}\left[ {\mathbf{Z}(t) \atop \mathbf{H}(t)} \right] = \left[ f(\mathbf{Z}(t); \mathbf{\Theta})g(\mathbf{H}(t); \mathbf{\Phi})\frac{d\Tilde{\mathbf{X}}(t)}{dt} \atop g(\mathbf{H}(t); \mathbf{\Phi})\frac{d\Tilde{\mathbf{X}}(t)}{dt} \right],
\end{equation}

where $\mathbf{H}(0) = \text{FC}\big(\Tilde{\mathbf{X}}{(t_0)}\big)$ and $\mathbf{Z}(0) = \text{FC}\big(\mathbf{H}(0)\big)$ are the initial values of the two NCDEs. We use $\mathbf{Z}(\tau)$ to conduct one-step-ahead forecast: $\Hat{\mathbf{y}} = \text{FC}\big( \mathbf{Z}(\tau) \big) \in \mathbb{R}^{N}$ where N is number of channels in \mts. Afterwards, we optimize the entire network using the loss function below:

\begin{equation}
\mathbf{L} = \frac{\sum_{i=1}^{N} m_{i}^{(t)} \cdot || y^{(t)}_{i} - \Hat{y}^{(t)}_{i} ||_1}{|N|},
\end{equation}
where $y_i $ and $\Hat{y}_i $ are the real and predicted value for node $i$ respectively. The $m_{i}^{(t)} \in \{0,1\}$ denotes whether the real values are not missing at timestamp $t$, where $m_{i}^{(t)} \in \mathbf{M}$ and $\mathbf{M} \in \{0,1\}^{N}$. The masked training approach prevents \ourmethod to be fitted on potentially noisy imputed values in training data, and forces \ourmethod to learn the \mts representation purely from what is available in the non-missing training observations.
Lastly, we can also use different ODE solvers to solve the augmented ODE, including the explicit Euler method, the 4th-order Runge-Kutta method, and the Dormand-Prince method \cite{chen2018neural}.

\section{Experimental Study}\label{sec:experiments}
In this section, we conduct experiments to explore capabilities of \ourmethod by answering the following questions: 
\begin{itemize} 
    \item \textbf{Irregular MTS Anomaly Detection} Does our framework outperform baseline methods in real-time \textit{irregular} \mts anomaly detection tasks? More importantly, does the performance remains stable with the increase in missing rates?
    \item \textbf{Regular MTS Anomaly Detection} As a control setting, does our framework still outperform baseline methods in real-time \textit{regular} \mts anomaly detection tasks? 
    \item \textbf{Ablation Study} What are respective contributions of the specific modules in \ourmethod?
    \item \revision{\textbf{Robustness Analysis} What extent of missing rate would notably impact GST-Pro's performance?}
\end{itemize}

\subsection{Experimental Settings} \label{subsec:exp_setting} 
In this subsection, we introduce the experimental settings to empirically evaluate our approach against state-of-the-art methods on real-world datasets at increasing missing rates. 

\subsubsection{Datasets}
We evaluate \ourmethod on two widely used realistic datasets for multivariate time series anomaly detection: SWaT and WADI. Both datasets are water treatment physical test-bed systems with simulated attack scenarios based on real-world water treatment plants. The statistics of these datasets are demonstrated in Table \ref{table:dataset}, and the detailed descriptions are given as follows:

\begin{table}[!hbt]
	\centering
	\caption{The statistics of the datasets.}
	\begin{tabular}{@{}c|c|c|c|c@{}}
		\toprule
		\textbf{Dataset} &  \textbf{$\sharp$ channels} & \textbf{$\sharp$ train} & \textbf{$\sharp$ test} &\textbf{anomalies} \\ \midrule
		\textbf{SWaT}   & 51    & 47,515     & 44,986            & 11.97\%              \\
		\textbf{WADI}             & 127    & 118,795     & 17,275         & 5.99\%               \\
		\bottomrule
	\end{tabular}
	\label{table:dataset}
\end{table}
\begin{itemize} 
	\item \textbf{SWaT\cite{mathur2016swat}} Secure WAter Treatment dataset has a training set with 7 days of operations that is non-anomalous and 4 days of the test set with multiple realistic simulated attack scenarios. The attacks that are conducted at different intervals in the test set are the positive anomaly labels that represent the attacks, while the rest of the timestamps are labelled as negatives. SWAT is a scaled-down real-world industrial water treatment plant dataset initiated by Singapore’s Public Utility Board, making it a realistic test bed for the empirical evaluation of models. 
 
    \item \textbf{WADI\cite{ahmed2017wadi}} WADI extends SWAT by having a larger number of pipelines, storage, and treatment systems. The scale of the WADI better represents a realistic water treatment dataset \cite{deng2021graph}. The train set has two weeks of non-anomalous data while the test set lasts for 2 days with multiple attacks conducted at different intervals. 
 \end{itemize}
Following the implementation of original author~\cite{deng2021graph}, we removed the first 21,600 samples and down-sampled SWaT and WADI to one measurement every 10 seconds by taking the median values. We keep the last 10\% of the training data as the validation set. To obtain irregular data with missing values, we randomly generate a masking series, $\{m_i^{(1)}, m_i^{(2)},\cdots,m_i^{(T)}\}$ to drop the real observations at a pre-defined missing rate for each channel or node independently. For each channel, its sensing observation at timestamp, $t$, is said to be missing if $m^{(t)}_{i} = 0$, and available if $m^{(t)}_{i} = 1$. 

We generate three settings by randomly dropping 10\%, 30\%, and 50\% of observed values. To ensure reproducibility, the same mask is generated as a pre-processing step on the \mts data, and the same data is used to evaluate the baseline and \ourmethod performances. For each experimental seed, we also generate a new masking series to ensure all models are fairly and comprehensively assessed. 

\subsubsection{Baselines}
As state-of-the-art methods require real observations, we first impute the missing values in the dataset using three standard imputation approaches:
\begin{itemize} 
    \item \textbf{Naive Imputation} For each sensor, the naive imputation method replaces the missing values in the data using the most recent available value.
    \item \textbf{Mean Imputation} For each sensor, the mean imputation method ignores the missing data to compute the mean value in the training data. 
    \item \textbf{Cubic Spline Imputation} The cubic spline imputation first generates a sliding window input. Then, it performs interpolation within the windows to fill up the missing values by treating the window values as the boundaries of a set of known points to avoid future information leakage. 
 \end{itemize}

After imputing missing values, baselines are implemented following the recommendations of original authors.
\begin{itemize} 
    \item \textbf{LSTM-VAE \cite{park2018multimodal}} is a reconstruction approach that assesses each timestamp abnormality based on the reconstruction errors. Variational Autoencoder (VAE) is applied to model the underlying probability distribution of the \mts values, while the LSTM module replaces the feed-forward neural networks in the original VAE to capture the temporal depencencies of \mts. 
    \item \textbf{OmniAnomaly~\cite{su2019robust}} 
     Similar to LTSM-VAE, OmniAnomaly is also a reconstruction approach. Using a stochastic recurrent neural network and planar normalizing flow, OmniAnomaly explicitly models the temporal dependencies to generate the reconstruction probabilities of the current observation. The anomaly score is the posterior reconstruction probability of each input.  
    \item \textbf{GDN \cite{deng2021graph}} is an attention-based GNN forecasting method. It learns spatial relationships between multivariate channels for one-step-ahead forecasting and determines anomaly scores based on maximum forecast deviations. The original model needs a full test set for error normalization at each timestamp, making it suitable for aftermath detection rather than real-time detection model. We modify this by using the validation set median, enabling real-time anomaly detection for GDN.
 \end{itemize}
\subsubsection{Parameter Settings} The parameterized fully-connection layers $\text{FC}(\cdot)$ for the spatial and temporal process have three hidden layers with the hidden dimension being set as 128. We train our model for 100 epochs with early stopping of 15 epochs. We use a batch size of 64 and Adam optimizer is applied to optimize \ourmethod with learning rate of 0.001, $(\beta_{1},\beta_{2}) = (0.9,0.999)$, and weight decay of 0.001. We also clip global norm of the gradient at 5.0. Validation set ratio for SWaT and WADI are fixed at 0.1 across all experiments. We set sliding window length, $w_s$, to be 5 for SWaT and WADI, as suggested by the original author \cite{deng2021graph}. The \ourmethod's forecasting module. Finally, the anomaly scorer only has a single parameter, $W$, which is set to be 50,000 or the maximum timestamps of the dataset for all settings.  

\subsubsection{Computing Infrastructures}
We performed all tests on a personal computer running Ubuntu 20.04, with an NVIDIA Tesla T4 GPU, a 2.20GHz Intel Xeon CPU, and 12.7 GB RAM. We used seeds 1-5 to mask dataset values and for model comparisons across five runs. \revision{Given the critical importance of efficiency in real-time anomaly detection, we assessed the average inference times per time point for SWaT and WADI. The results were as follows: \ourmethod required approximately 0.92 milliseconds for SWaT and 1.2 milliseconds for WADI. In comparison, OmniAnomaly recorded times of 0.18 milliseconds for SWaT and 0.25 milliseconds for WADI, LSTM-VAE took 0.2 milliseconds for SWaT and 0.5 milliseconds for WADI, and GDN needed 0.4 milliseconds for SWaT and 0.8 milliseconds for WADI.}

\subsection{Anomaly Detection Results}
As anomaly thresholds can vary based on applications, we follow previous works \cite{zong2018deep,park2018multimodal,li2021multivariate} in measuring anomaly detection performance using scale-invariant metrics that do not require thresholds. For each timestamp, a model has to be correctly labeled the timestamp as anomalous or not \cite{deng2021graph}. Hence, the closer the ROC and PRC score is to 1, the better a model is at providing a useful anomaly indicator that clearly differentiates every anomalous and non-anomalous time point.
\begin{table*}[!htb]
\scriptsize
\centering
\caption{Average AUC performance ($\pm$ standard deviation) of five experimental runs on SWaT benchmark datasets. \\ 10\%, 30\% and 50\% represent the missing rates of the datasets. AUC values are rescaled to 0-100.}

\label{tab:swat_irregular}
\resizebox{\textwidth}{!}{
\begin{tabular}{cc|ccc|ccc}
\toprule
\multicolumn{1}{c}{\multirow{2}{*}{\begin{tabular}[c]{@{}l@{}}\\\end{tabular}}}  & \multirow{2}{*}{Methods} & \multicolumn{3}{c|}{ROC-AUC} & \multicolumn{3}{c}{PRC-AUC} \\ 
\multicolumn{1}{l}{} &  & 10\% & 30\% & 50\% & 10\% & 30\% & 50\% \\ \midrule
\multirow{3}{*}{{\begin{tabular}[c]{@{}l@{}}\;\;\; Naive \\ \;Imputation\end{tabular}}} & LSTM-VAE & 79.6 $\pm$ 0.4 & 79.6 $\pm$ 0.4 & 78.9 $\pm$ 2.1 & 68.6 $\pm$ 0.8  & 67.7 $\pm$ 3.6 & 58.9 $\pm$ 13.3 \\
 & OmniAnomaly & 82.9 $\pm$ 0.4 & 80.0 $\pm$ 3.8	&	79.9 $\pm$ 4.3 & 73.1 $\pm$ 2.6 &	56.7 $\pm$ 20.9 & 56.1 $\pm$ 24.1 \\
 & GDN-GNN & 70.8 $\pm$ 18.5 & 71.0 $\pm$ 17.5 & 69.9 $\pm$ 16.6 & 49.3 $\pm$ 33.8 & 49.9 $\pm$ 34.2 & 48.3 $\pm$ 32.5  \\ \midrule
\multirow{3}{*}{{\begin{tabular}[c]{@{}l@{}}\;\;\; Mean \\ \;Imputation\end{tabular}}} & LSTM-VAE & 80.2 $\pm$ 1.5 & 79.9 $\pm$ 0.4 & 79.9 $\pm$ 0.4 & 69.1 $\pm$ 0.4 & 67.2 $\pm$ 2.1 & 64.8 $\pm$ 3.6 \\
 & OmniAnomaly & 81.6 $\pm$ 0.8  &	81.6 $\pm$ 0.8 &	80.9 $\pm$ 0.7 &	67.4 $\pm$ 3.5 &	67.8 $\pm$ 6.1 &	63.2 $\pm$ 5.4 \\
 & GDN-GNN & 71.7 $\pm$ 15.6 & 77.2 $\pm$ 11.5 & 76.1 $\pm$ 11.4 & 47.7 $\pm$ 32.3 & 57.2 $\pm$ 25.3 & 45.8 $\pm$ 21.6 \\
\midrule
\multirow{3}{*}{{\begin{tabular}[c]{@{}l@{}}Cubic Spline \\ \;\;Imputation\end{tabular}}} & LTSM-VAE & 79.9 $\pm$ 0.3 & 79.4 $\pm$ 0.9 & 79.6 $\pm$ 0.5 & 68.9 $\pm$ 0.4 & 64.7 $\pm$ 8.6 & \textbf{68.6 $\pm$ 0.9}  \\
 & OmniAnomaly & 82.1 $\pm$ 0.6 &	74.4 $\pm$ 0.4 & 74.0 $\pm$ 0.4 &	69.6 $\pm$ 0.7 &	26.2 $\pm$ 2.8 & 26.6	$\pm$ 2.4 \\
 & GDN-GNN & 69.7 $\pm$ 17.0 & 68.6 $\pm$ 15.5 & 63.5 $\pm$ 16.8 & 47.0 $\pm$ 31.9 & 48.4 $\pm$ 33.0 & 35.5 $\pm$  32.2 \\
\midrule 
& \ourmethod & \textbf{85.5 $\pm$ 0.3} & \textbf{86.3 $\pm$ 0.3} & \textbf{86.2 $\pm$ 0.7} &  \textbf{73.3 $\pm$ 0.6} & \textbf{69.0 $\pm$ 0.4} & 67.0 $\pm$ 1.6
 \\ 
 
 \bottomrule
\end{tabular}
}

\end{table*}

\begin{table*}[!htb]
\scriptsize
\centering
\caption{Average AUC performance ($\pm$ standard deviation) of five experimental runs on WaDI benchmark datasets. \\ 10\%, 30\% and 50\% represent the missing rates of the dataset. AUC values are rescaled to 0-100.}

\label{tab:wadi_irregular}
\resizebox{\textwidth}{!}{
\begin{tabular}{cc|ccc|ccc}
\toprule
\multicolumn{1}{c}{\multirow{2}{*}{\begin{tabular}[c]{@{}l@{}}\\\end{tabular}}}  & \multirow{2}{*}{Methods} & \multicolumn{3}{c|}{ROC-AUC} & \multicolumn{3}{c}{PRC-AUC} \\ 
\multicolumn{1}{l}{} &  & 10\% & 30\% & 50\% & 10\% & 30\% & 50\% \\ \midrule
\multirow{3}{*}{{\begin{tabular}[c]{@{}l@{}}\;\;\; Naive \\ \;Imputation\end{tabular}}} & LSTM-VAE & 50.3 $\pm$ 6.5 & 50.2 $\pm$ 4.3 & 49.4 $\pm$ 3.2 & 14.3 $\pm$ 8.8 & 13.4 $\pm$ 8.8 & 16.0 $\pm$ 3.4 \\
 & OmniAnomaly & 54.7 $\pm$ 0.8 & 54.1 $\pm$ 0.8	& 54.1 $\pm$ 2.5	&	22.1 $\pm$ 1.5 &	17.2 $\pm$ 5.5 & 19.6 $\pm$ 2.7 \\
 & GDN-GNN & 48.4 $\pm$ 1.2 & 48.4 $\pm$ 1.5 & 48.1 $\pm$ 1.1 & 6.3 $\pm$ 1.2 & 5.0 $\pm$ 0.1 & 6.0 $\pm$ 1.4 \\ \midrule
\multirow{3}{*}{{\begin{tabular}[c]{@{}l@{}}\;\;\; Mean \\ \;Imputation\end{tabular}}} & LSTM-VAE & 47.0 $\pm$ 3.8 & 48.5 $\pm$ 2.0 & 49.6 $\pm$ 1.3 & 7.9 $\pm$ 5.3 & 7.2 $\pm$ 2.0 &  7.1 $\pm$ 2.7 \\
 & OmniAnomaly & 56.3 $\pm$ 1.4 &	60.5 $\pm$ 0.8 & 63.6 $\pm$ 14.0	&	19.8 $\pm$ 1.2 &	17.1 $\pm$ 1.5 &	14.0 $\pm$ 1.2 \\
 & GDN-GNN & 48.4 $\pm$ 0.4 & 48.6 $\pm$ 0.6  & 49.5 $\pm$ 0.4 & 5.4 $\pm$ 0.6 & 5.3 $\pm$ 0.1 & 5.4 $\pm$ 0.1 \\
\midrule
\multirow{3}{*}{{\begin{tabular}[c]{@{}l@{}}Cubic Spline \\ \;\;Imputation\end{tabular}}} & LTSM-VAE & 47.9 $\pm$ 2.6 & 50.5 $\pm$ 2.1 & 49.7 $\pm$ 1.9 & 11.9 $\pm$ 5.1 & 13.0 $\pm$ 5.2 & 11.2 $\pm$ 0.4 \\
 & OmniAnomaly & 52.7 $\pm$ 1.2 & 53.1 $\pm$ 1.2	& 56.0 $\pm$ 4.7 &	17.7 $\pm$ 0.7 &	15.9 $\pm$ 5.1 &	18.3 $\pm$ 0.6 \\
 & GDN-GNN & 48.0 $\pm$ 0.5 & 47.1 $\pm$ 0.5 & 45.9 $\pm$ 0.9 & 5.4 $\pm$ 0.5 & 5.3 $\pm$ 0.4 & 5.3 $\pm$ 0.4 \\
\midrule 
& \ourmethod & \textbf{73.9 $\pm$ 0.6}  & \textbf{74.2 $\pm$ 0.4} & \textbf{72.6 $\pm$ 0.4} & \textbf{37.3 $\pm$ 0.2} & \textbf{37.0 $\pm$ 0.6} & \textbf{34.8 $\pm$ 0.6}
 \\ 
 \bottomrule
\end{tabular}
}
\end{table*}

\subsubsection{Irregular MTS Anomaly Detection} From the ROC and PRC results in Table \ref{tab:swat_irregular} and \ref{tab:wadi_irregular}, we notably observed that:
\begin{itemize} 
    \item \textbf{Performance of Proposed Framework} \ourmethod outperforms the baselines across all settings, except for SWaT with the missing rate of 50\% where \ourmethod closely matches the performance of LSTM-VAE using cubic spline imputation. On average, we outperform the second-best baseline by 13.91\% and 46.66\% on ROC-AUC and PRC-AUC scores respectively. Notably, we outperform the second-best method, Naive-OmniAnomaly, on WADI's PRC-AUC results by 68\% to 115\%.
    \item \textbf{Robustness to Increase in Missing Rates} \ourmethod shows not only an overall outperformance but achieves the lowest variability in anomaly detection performance with the change in missing rates. This results in a widening performance gap between \ourmethod and the baselines as the missing rate increases. LSTM-VAE and OmniAnomaly that take the reconstruction approach also achieve fairly stable performances, except for OmniAnomaly using Cubic Spline Imputation. On the other hand, as compared to its performance under regular \mts settings, GDN-GNN shows a significant deterioration in performance under irregular time series settings. As GDN-GNN takes the maximum deviation between the predicted and observed values among the \mts channels, we hypothesize that this mechanism is sensitive to a small noise in the dataset values. This undesirable property is exacerbated under high missing rate scenarios because the imperfection in the missing data imputation can materially disrupt its anomaly detection performances.
    \item\textbf{Modular Pipeline leads to Performance Variability} \revision{In our analysis of baseline performances, it becomes evident that no single method of data imputation stands out as universally superior. The effectiveness of these methods is inherently tied to (i) the specific characteristics of the datasets and (ii) the type of model employed. For instance, while mean imputation method shows promising results in the SWaT dataset, naive imputation method appears more effective in the context of WADI. Similarly, the performance of LSTM-VAE with cubic spline imputation in SWaT highlights the variability in effectiveness among different combinations of datasets and baseline methods. In contrast, \ourmethod illustrates a more robust and adaptive approach. By learning from non-missing observations in the training data, \ourmethod captured the spatiotemporal dependencies effectively directly from the non-missing data. This process, significantly enhanced by the integration of the NCDE model, steers clear of relying on predefined assumptions about data characteristics. Such an approach not only increases the robustness and accuracy of anomaly detection but also showcases \ourmethod's adaptability to diverse datasets.}
\end{itemize}
\revision{The cornerstone of \ourmethod lies in its adept handling of the complexities inherent in real-world data, a facet where traditional methods, reliant on static impute-then-detect techniques, often falter due to their limited capacity to capture the nature of datasets. Both the baselines and \ourmethod can learn from spatiotemporal patterns. However, only \ourmethod can ensure a nuanced and accurate representation of underlying processes directly from time series that is marred by missing values due to irregular sampling or partial observation.}

\revision{Here, \ourmethod's NCDEs module plays a pivotal role, modeling the hidden state dynamics as controlled differential equations, thereby maintaining the temporal data's integrity and capturing complex, evolving relationships. This unified approach of \ourmethod is not just a feature but a necessity. It addresses the intertwined challenges of imputing missing values, making accurate forecasts and detecting outliers. This strategy is crucial as separating these tasks can lead to significant loss of temporal information and a failure to accurately represent continuous-time processes. Overall, \ourmethod's unified strategy sets a new benchmark, outperforming methods that treat these problems in isolation.}

\subsubsection{Regular MTS Anomaly Detection}
While we focus on \mts anomaly detection with missing values, it is equally important that \ourmethod can perform well under regular \mts anomaly detection. This empirically evaluates if \ourmethod can capture the spatiotemporal dependencies of the data, and hence output anomaly indicators that can alert operators that anomalous events have (or have not) occurred. More importantly, it also stress tests the validity of our assumption that current timestamp values are not required to score anomalies.
\begin{figure*} [!hbt]
    \centering \includegraphics[width=.8\textwidth]{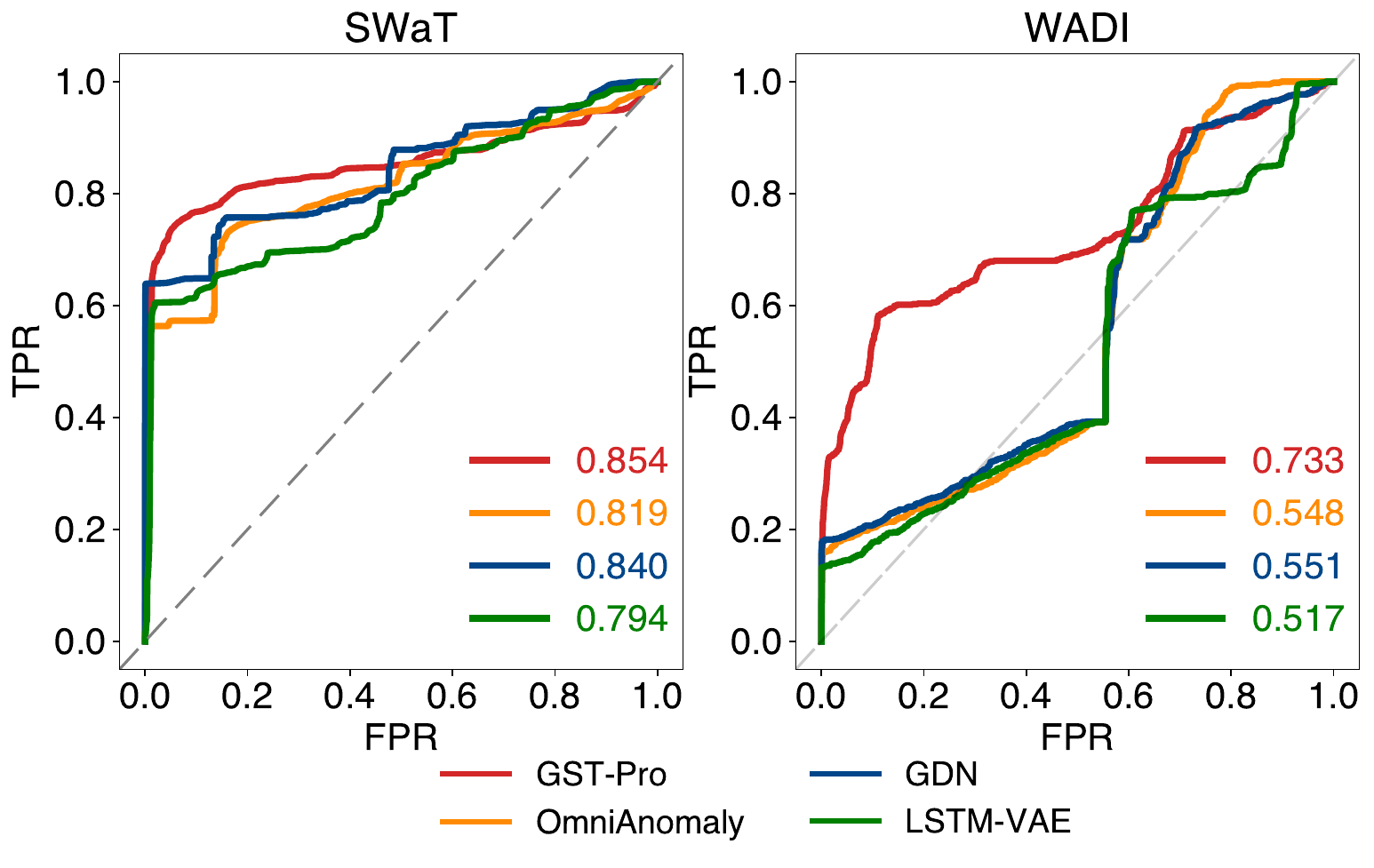}\\
    \caption{ROC-AUC performances on SWaT and WADI. TPR and FPR represents True and False Positive Rate. \ourmethod achieves state-of-the-art results with ROC-AUC 0.854 and 0.733 on SWaT and WADI.}
    \label{fig. roc_auc}
\end{figure*}
As shown in Figure \ref{fig. roc_auc}, \ourmethod outperforms the baselines under the regular setting, suggesting computing forecasting or reconstruction errors are \textit{not} required to detect anomalies. This approach deviates from recent state-of-the-art models \cite{garg2021evaluation,darban2022deep,schmidl2022anomaly}, and illustrates an effective alternative to detecting anomalies for \mts data. In the irregular settings, we have also demonstrated that this is a pragmatic approach as \ourmethod only requires \textit{some} real observations in the sliding window to make a one-step-ahead forecast and \textit{no} real observation at the current timestamp to detect anomalies accurately under high missing rate settings.
\subsection{Ablation Study}
We conduct an ablation study on SWaT (Table \ref{tab:swat_ablation}) and WADI (Table \ref{tab:wadi_ablation}) with 0\% and 50\% missing rate to validate how various modules of \ourmethod contribute to its irregular multivariate time series anomaly detection performance. We modify two major modules of \ourmethod, the forecasting module (FM) and the Anomaly Detection Module (AD). For the FM module, we implement made the following modifications: 

\begin{itemize} 
    \item \textbf{w/o SP}
    \ourmethod without the Spatial Process module is achieved by removing graph convolution layer and the learned adjacency matrix.
    \item \textbf{w/o TP} \ourmethod without the Temporal Process module entirely relies on the Spatial process module to implicitly model the temporal dependencies.
\end{itemize}
For AD module modifications, we keep the forecasting module fixed and replace the gaussian scorer with the Principal component analysis (PCA) scorer, or a Kmeans scorer. The former represents a reconstruction-based approach while the latter represents a distance-based approach. For both scorers, we normalize the forecast deviation and forecast values using the median and inter-quartile range \cite{deng2021graph}. This is to dampen the small spikes in forecast values even when the system behavior is not anomalous. The scorers are fitted on the validation set and evaluated on the test set:
\begin{itemize} 
    \item \textbf{ReconPCA} For each timestamp with non-missing observations, we initially compute the forecasting deviation in order to lessen the disparity in characteristics among the variable channels. Following this, we implement Probabilistic PCA \cite{tipping1999probabilistic} to calculate the average reconstruction errors of the non-missing forecasting deviations for each timestamp. The average reconstruction error at each timestamp is treated as the indicator for anomalies.
    \item \textbf{DistKmeans} For the validation set, we first apply Kmeans to generate multiple clusters. The number of cluster, K, is determined by using Silhouette score\cite{rousseeuw1987silhouettes} and we search K from 0 to 20. Following this, we calculate the distance between forecast and the centroid of its closest corresponding cluster. The computed distance is used as the anomaly score for detecting anomalies.
\end{itemize}

\begin{table}[!hbt]

\caption{Ablation Study - Average AUC performance on SWaT with 0\% and 50\% missing rate. AUC values are rescaled to 0-100.}
\label{tab:swat_ablation}
\centering
\resizebox{7.7cm}{!}{
\begin{tabular}{cc|cc|cc}
\toprule
\multicolumn{1}{c}{\multirow{2}{*}{\begin{tabular}[c]{@{}l@{}}\\\end{tabular}}}  & \multirow{2}{*}{Methods} & \multicolumn{2}{c|}{ROC-AUC} & \multicolumn{2}{c}{PRC-AUC} \\ 
\multicolumn{1}{l}{} &  & 0\% & 50\% & 0\% & 50\% \\ \midrule

& \textbf{\ourmethod} & \textbf{85.54} & \textbf{86.21} & \textbf{73.31} & \textbf{66.96} \\ \midrule
 \multirow{2}{*}
{{\begin{tabular}[c]{@{}l@{}} \textbf{FM}\end{tabular}}} & \textbf{w/o SP} & 83.70 & 82.74 & 63.76 & 55.83  \\
 & \textbf{w/o TP} & 83.55 & 81.52 & 63.71 & 54.64  \\
 \midrule
\multirow{2}{*}{{\begin{tabular}[c]{@{}l@{}} \textbf{AD} \end{tabular}}} & \textbf{ReconPCA} & 83.29 & 79.11 & 59.69 & 44.35 \\
 & \textbf{DistKmeans} & 77.88  &	77.03 &	67.36 & 65.61 \\
 \bottomrule
\end{tabular}
}

\end{table}

\begin{table}[!hbt]

\caption{Ablation Study - Average AUC performance on WADI with 0\% and 50\% missing rate. AUC values are rescaled to 0-100.}
\label{tab:wadi_ablation}
\centering
\resizebox{7.7cm}{!}{
\begin{tabular}{cc|cc|cc}
\toprule
\multicolumn{1}{c}{\multirow{2}{*}{\begin{tabular}[c]{@{}l@{}}\\\end{tabular}}}  & \multirow{2}{*}{Methods} & \multicolumn{2}{c|}{ROC-AUC} & \multicolumn{2}{c}{PRC-AUC} \\ 
\multicolumn{1}{l}{} &  & 0\% & 50\% & 0\% & 50\% \\ \midrule

& \textbf{\ourmethod} & \textbf{73.34} & \textbf{72.64} & \textbf{37.21} & \textbf{34.80} \\ \midrule
 \multirow{2}{*}
{{\begin{tabular}[c]{@{}l@{}} \textbf{FM}\end{tabular}}} & \textbf{w/o SP} & 73.29 & 71.26 & 27.02 & 25.82 \\
 & \textbf{w/o TP} & 73.54 & 71.22 & 27.08 & 25.64  \\
 \midrule
\multirow{2}{*}{{\begin{tabular}[c]{@{}l@{}} \textbf{AD} \end{tabular}}} & \textbf{ReconPCA} & 64.83 & 62.14 & 19.79 & 15.74 \\
 & \textbf{DistKmeans} & 53.68  &	54.86 &	16.72 & 14.77 \\
 \bottomrule
\end{tabular}
}
\end{table}

From Tables \ref{tab:swat_ablation} and \ref{tab:wadi_ablation}, we observe a significant drop PRC-AUC performance upon removal of the spatial or temporal module. This strongly supports our hypothesis that modeling spatial-temporal dependencies is critical for multivariate anomaly detection. Secondly, the PCA reconstruction approach shows a greater decline in performance when the \mts encounters irregularity. As we argued earlier, we propose that dependence on timepoint-detection approaches can result in unstable performance due to inherent noise, unpredictability, and unreliability of real observed values during anomalous periods. Lastly, Kmeans fails to yield satisfactory results as it treats each multivariate observation as an independent sample instead of a sequence in a temporal series.

\subsection{Robustness Analysis}
\begin{figure*} [!hbt]
    \centering \includegraphics[width=.8\textwidth]{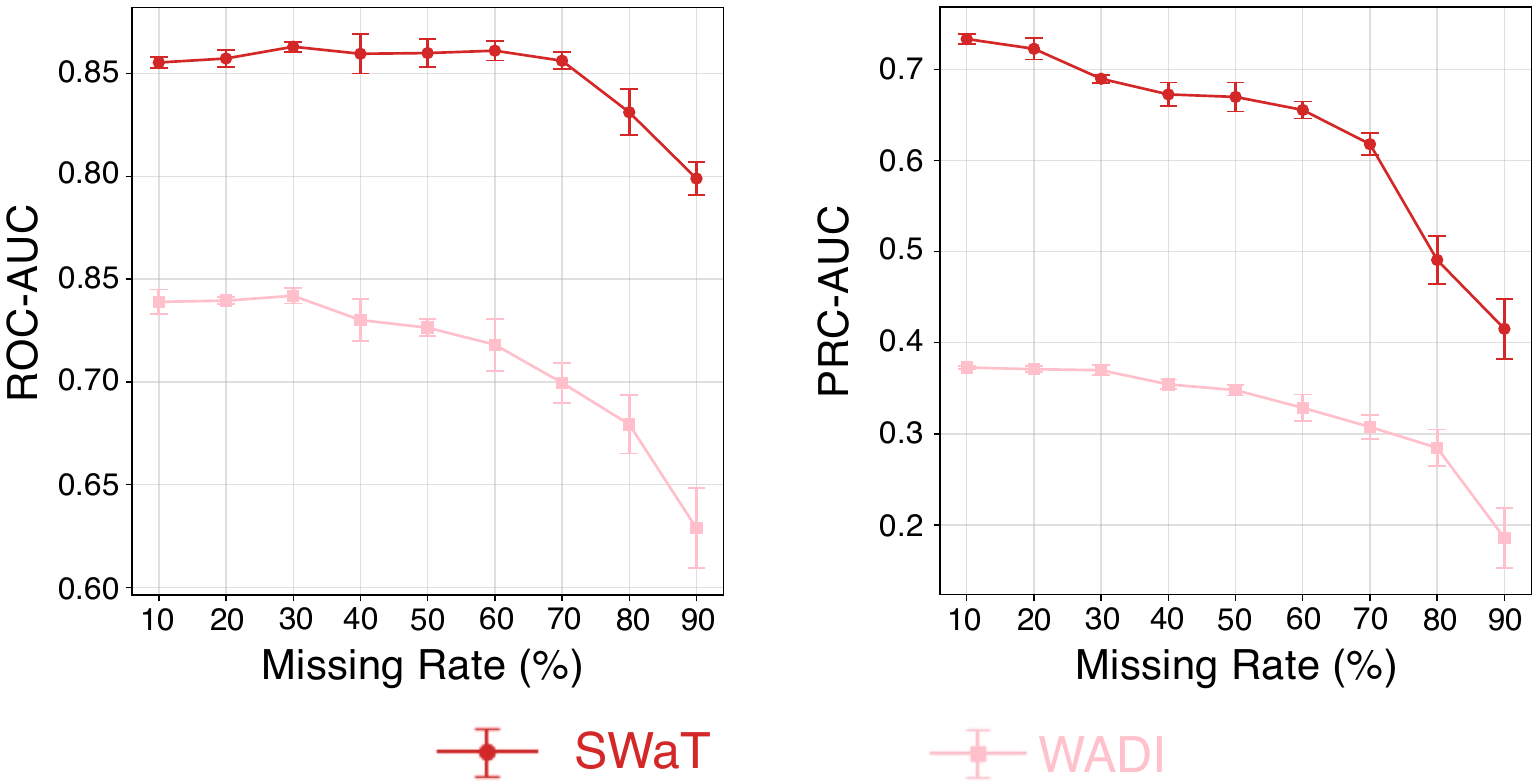}\\
    \caption{ROC-AUC and PRC-AUC performances on SWaT and WADI from 10\% to 90\% missing rate. GST-Pro still achieves state-of-the-art ROC-AUC results at 70\% missing rate on SWaT (0.856) and 90\% on WADI (0.629), as compared to the best competing baseline on regular MTS Anomaly Detection setting in Figure \ref{fig. roc_auc}.}
    \label{fig. robust}
\end{figure*}

\revision{In this subsection, we delve into how \ourmethod copes with varying levels of missing data, ranging from 10\% to 90\%. The analysis, as illustrated in Figure \ref{fig. robust}, showcases the model's performance in terms of ROC-AUC and PRC-AUC on the SWaT and WADI datasets under these varying conditions. Remarkably, \ourmethod maintains its state-of-the-art performance even at high missing rates, achieving a ROC-AUC of 0.856 on SWaT with a 70\% missing rate and 0.629 on WADI with a 90\% missing rate. This performance is particularly notable since \ourmethod still outperforms the best-performing baseline models in a regular MTS Anomaly Detection setting, even under these high missing rate scenarios.}

\revision{Overall, these results indicate that \ourmethod's performance begins to be notably impacted only at extremely high missing rates. It is specifically beyond the 70\% threshold where the performance decrement for \ourmethod becomes more pronounced. Intriguingly, \ourmethod can even outperform the competing baselines in regular scenarios when under extreme missing value conditions. This analysis not only underscores \ourmethod's resilience in handling incomplete data but also assists in determining the thresholds beyond which missing data significantly affects its accuracy and reliability.}

\revision{Moving forward, we intend to further advancing the scalability of \ourmethod, ensuring its robust applicability in real-world scenarios. Our ongoing efforts will concentrate on optimizing the model's architecture and algorithms to efficiently handle even larger datasets, a crucial step towards broadening its practical utility. Particularly, we seek to leverage the ability of NCDE~\cite{chen2018neural} in \ourmethod model for its capacity to use memory-efficient training with adjoint-based backpropagation~\cite{kidger2020neural}. This approach is akin to that used in invertible networks~\cite{behrmann2019invertible}. This progression should enable \ourmethod to learn from diverse data patterns and structures, effectively handling the intricacies of larger and more diverse datasets.}

\revision{Additionally, our goal is to expand \ourmethod's applicability across a wider spectrum of uses by tackling concept drift. We seek to incorporate methods to identify and measure data drift. This will enable timely modifications to the model to align with changing data distributions~\cite{webb2016characterizing,goldenberg2020pca}. This enhancement not only aims to improve the model's adaptability but also ensures its relevance in dynamic data environments.}

\section{Conclusion}\label{sec:conclusion}
In this work, we propose a novel framework to address irregular \mtsad. Our model, \ourmethod, robustly detects anomalies in real-time even when current observations are completely absent. Experiments on real-world datasets showed that \ourmethod not only outperformed state-of-the-art baselines in regular \mts anomaly detection settings but also in irregular \mts with high missing rate scenarios, paving the way for deep STGNN methods to be implemented in real-world applications. In the future, we will exploit the integration of graphs and large language models (LLMs) \cite{pan2023integrating, pan2023unifying, luo2023reasoning} for effective representation learning for time series; alternatively, we will directly reprogram LLMs for time series anomaly detection \cite{jin2023time,jin2023large}.



\bibliographystyle{elsarticle-num} 
\bibliography{reference}






\end{document}